\documentclass[runningheads]{llncs}

\usepackage{booktabs} 
\usepackage{graphicx}
\usepackage[russian,english]{babel}

\usepackage[utf8]{inputenc}
\usepackage[T1,T2A]{fontenc}
\usepackage{multicol}
\usepackage{tabularx,colortbl}
\usepackage{makecell}
\usepackage{multirow}
\usepackage{rotating}

\usepackage{amsmath}
\usepackage{amssymb}

\usepackage{graphicx}
\begin{document}
\title{Char-RNN and Active Learning for Hashtag Segmentation}
\titlerunning{Char-RNN for Hashtag Segmentation}
%
\author{Taisiya Glushkova \and Ekaterina Artemova}
\authorrunning{T. Glushkova \and E. Artemova}

\institute{National Research University Higher School of Economics  \\
20 Myasnitskaya Ulitsa, Moscow 101000, Russia \\
\email{toglushkova@edu.hse.ru,echernyak@hse.ru}}

\maketitle              
\begin{abstract}

We explore the abilities of character recurrent neural network (char-RNN) for hashtag segmentation. Our approach to the task is the following: we generate synthetic training dataset according to frequent  $n$-grams that satisfy predefined morpho-syntactic patterns to avoid any manual annotation. The active learning strategy limits the training dataset and selects informative training subset. The approach does not require any language-specific settings and is compared for two languages, which differ in inflection degree. 

\keywords{Hashtag segmentation \and Recurrent neural network \and Character level model}
\end{abstract}

\section{Introduction}

A hashtag is a form of metadata labeling used in various social networks to help the users to navigate through the content. For example, one of the most popular hashtags on Instagram is "\#photooftheday" [photo of the day]. Hashtags are written without any delimiters, although some users use an underscore or camel-casing to separate words. Hashtags themselves may be a great source for features for following opinion mining and social network analysis. Basically hashtags serve as keyphrases for a post in social media. By segmenting the hashtags into separate words we may use regular techniques to process them. The problem of hashtag segmentation resembles of another problem, namely word segmentation.    

The problem of word segmentation is widely studied in languages like Chinese, since it lacks whitespaces to separate  words, or in German to split compound words. In languages like English or Russian, where compounds are not that frequent as in German and where whitespace delimiters are regularly used, the problem of word segmentation arises mainly when working with hashtags. 

Formally the problem is stated as follows: given a string of $n$ character $s = s_1 \ldots s_n$ we need to define the boundaries of the substrings $s_{i:j}, i < j$, so that each substring is meaningful (i.e. is a regular word, named entity, abbreviation, number, etc). The main challenge of this problem is that the segmentation might be ambiguous. For example, a string ``somethingsunclear'' might be segmented as ``something sun clear'' or ``somethings unclear''. To deal with the ambiguity more processing is required, such as POS-tagging, estimation of frequencies of all hashtag constituencies or their co-occurence frequency. The frequencies can be estimated on a large corpus, such as BNC \footnote{https://corpus.byu.edu/bnc/}, COCA \footnote{https://corpus.byu.edu/coca/}, Wikipedia. However when working with noisy user generated data, such as texts or hashtags from social networks, the problem of unknown words (or out of vocabulary words) arises. In language modeling this problem is solved by using smoothing, such as Laplacian smoothing or Knesser-Ney smoothing. Otherwise additional heuristics can be used to extend the dictionary with word-like sequences of characters.
Unlike language modelling, in hashtag segmentation frequency estimation is not only source for defining word boundaries. Otherwise candidate substrings can be evaluated according to length \cite{bansal2015towards}. 

Several research groups have shown that introducing character level into models help to deal with unknown words in various NLP tasks, such as text classification \cite{joulin2017bag}, named entity recognition \cite{ma2016end}, POS-tagging \cite{santos2014learning}, dependency parsing \cite{alberti2017syntaxnet}, word tokenization and sentence segmentation \cite{shao2018universal} or machine translation \cite{chungcharacter,vaswani2017attention}. The character level model is a model which either treats the text as a sequence  of characters without any tokenization or incorporates character level information into word level information. Character level models are able to capture morphological patterns, such as prefixes and suffixes, so that the model is able to define the POS tag or NE class of an unknown word. 

Following this intuition, we use a character level model for hashtag segmentation. Our main motivation is the following: if the character level model is able to capture word ending patterns, it should also be able to capture the word boundary patterns. We apply a character level model, specifically, a recurrent neural network, referred further as char-RNN, to the task of hashtag segmentation. The char-RNN is trained and tested on the synthetic data, which was generated from texts, collected from social networks in English and Russian, independently. We generate synthetic data for training by extracting frequent $N$-grams and removing whitespaces. The test data is annotated manually \footnote{The test data is available at: \url{https://github.com/glushkovato/hashtag\_segmentation}}. Since the problem statement is very basic, we use additional techniques, such as active learning, character embeddings and RNN hidden state visualization, to interpret the weights, learned by char-RNN.  We address the following research questions and claim our respective contributions:

\begin{enumerate}
\item We show that our char-RNN model outperforms  the traditional unigram or bigram language models with extensive use of external sources~\cite{norvig2009natural,bansal2015towards}. 

\item What is the impact of high inflection in languages such as Russian on the performance of character-level modelling as opposed to  languages with little inflection such as English? We claim that character-level models offer benefits for processing highly inflected languages by capturing the rich variety of word boundary patterns. 

\item As getting sufficient amount of annotated training collection is labor-intensive and error-prone, a natural question would be: can we avoid annotating real-world data altogether and still obtain high quality hashtag segmentations? We approach this problem by using morpho-syntactic patterns to generate synthetic hashtags. 

\item A potentially unlimited volume of our synthetic training dataset raises yet another question of whether an informative training subset could be selected. To this extent, we apply an active learning-based strategy to subset selection and identify a small portion of the original synthetic training dataset, necessary to  obtain a high performance. 

\end{enumerate}
\section{Neural Model for Hashtag Segmentation}
\subsection{Sequence Labeling Approach}
We treat hashtag segmentation as a sequence labeling task. Each character is labeled with one of the labels $\mathcal{L} = \{0, 1\}$, (1) for the end of a word, and (0) otherwise (Table \ref{tbll} and \ref{tbl2}). Given a string $s = {s_1, \ldots, s_n}$ of characters, the task is to find the labels $Y^* = {y_1^*. \ldots, y_n^*}$, such that 
$ Y^* = \arg \max_{Y \in \mathcal{L} ^n} p(Y | s).$

\begin{table}[h!]
\centering
\caption{Illustration of sequence labeling for segmentation \#ремонтдома [``ремонт дома'' , house renovation]}
\label{tbll}
\begin{tabular}{|l|l|l|l|l|l|l|l|l|l|} \hline 
р & е & м & о & н & т & д & о & м & а \\ \hline 
0 & 0 & 0 & 0 & 0 & 1 & 0 & 0 & 0 & 1 \\ \hline
\end{tabular}

\vspace{3mm}
\caption{Illustration of sequence labeling for segmentation "\#photooftheday"  [photo of the day]}
\label{tbl2}
\begin{tabular}{|l|l|l|l|l|l|l|l|l|l|l|l|l|} \hline 
p & h & o & t & o & o & f & t & h & e & d & a & y \\ \hline 
0 & 0 & 0 & 0 & 1 & 0 & 1 & 0 & 0 & 1 & 0 & 0 & 1 \\ \hline
\end{tabular}
\end{table}
\vspace{-2mm}
\begin{figure}[h!]
\centering
\caption{Neural model for hashtag segmentation}
\includegraphics[width = 0.5\textwidth]{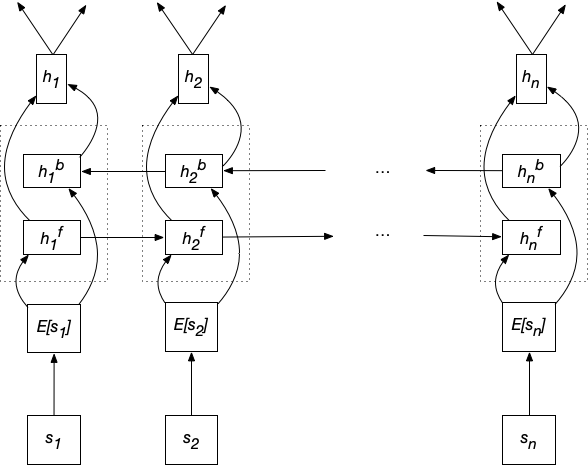}
\end{figure}
The neural model for hashtag segmentation consists of three layers. 
\begin{enumerate}
	\item The embedding layer is used to compute the distributed representation of input characters.  Each character $c_i$ is represented with an embedding vector $e_i \in \mathbb{R}^{d_e}$, where $d_e$ is the size of the character embedding. $E$ is the look up table of size $|V| \times d_e$, where $V$ is the vocabulary, i.e. the number of unique characters. 
	\item The feature layer is used to process the input. We use a bi-directional recurrent layer with LSTM units to process the input in forward and backward directions. The LSTM units we use are default keras LSTM units as introduced by Hochreiter.
	\item The inference layer is  used to predict the labels of each character. We use a single dense layer as f or inference and $softmax$ to predict the probabilities of the labels $\mathcal{L} = \{0, 1\}$. 
\end{enumerate}

\[softmax (y_{j} ) = \frac {e^{y_{j}}}{\sum _{k=1}^{|\mathcal{L}|} e^{y_{k}} }  \]

Each character is assigned with the most probable label.



The parameters of the char-RNN are the following:
\begin{enumerate}
    \item Embedding layer = 50 input dimensions;
    \item Feature layer = 64 bidirectional LSTM units;
    \item Inference layer = 2 output neurons with softmax activation function mapped to each of 64 outputs.
\end{enumerate}

\section{Dataset}

In this section we describe the datasets we used for hashtag segmentation. We experimented with Russian and English datasets to compare the performance of the char-RNN.

\subsection{Russian dataset}

To our knowledge there is no available dataset for hashtag segmentation in Russian, so we faced the need to create our own dataset. Our approach to the dataset creation was twofold: the training data was created from social network texts by selecting frequent $n$-grams and generating hashtags following some hashtag patterns. The test dataset consists of real hashtags collected from \url{vk.com} (a Russian social network) and were segmented manually.  

We followed the same strategy to create an English language dataset.

\subsubsection{Training Dataset Generation}

We scraped texts from several pages about civil services from  \url{vk.com}. Next we extracted frequent $n$-grams that do not contain stopwords and consist of words and digits in various combinations (such as word + 4 digits + word or word + word + 8 digits). We used several rules to merge these $n$-grams so that they resemble real hashtags, for example: 

\begin{itemize}
\item remove all whitespace: wordwordworddigits 

\textbf{Examples:} ЁлкаВЗазеркалье, нескольколетназад	
\item replace all whitespace with an underscore: word\_word\_digits

\textbf{Examples:} увд\_юга\_столицы
\item remove some whitespace and replace other spaces with an underscore: word\_worddigits.

\textbf{Examples:} ищусвоегогероя\_уфпс
\end{itemize}

A word here might be a word in lower case, upper case or capitalized or an abbreviation. There might be up to four digits.  

In general, we introduced 11 types of hashtags, which contain simply constructed hashtags as well as the complex ones. Here are a couple of examples:

\begin{itemize}
	\item   The hashtag consists of two parts: the word/abbreviation in the first part and the number or word in the second. The underscore is a delimiter.
    
	\textbf{Examples:} word\_2017, NASA\_2017, word\_word
    
	\item Two or three words, which are separated by an underscore.
    
	\textbf{Examples:} Word\_Word, word\_word\_word
\end{itemize}

\subsubsection{Test Dataset Annotation}
We segmented manually 2K the most frequent hashtags, extracted from the same collection of the scraped texts. 

The resulting size of the Russian dataset is 15k hashtags for training and 2k  hashtags for testing.  

\subsection{English dataset}

We used the dataset, released by \cite{bansal2015towards}. This dataset consists of: 
\begin{itemize}
\item a collection of tweets, which we used to generate the synthetic training hashtags according to the same rules as for Russian;
\item a collection of annotated and separated hashtags, which we used as a testing set. From this test set we excluded ambiguous hashtags, annotated with several possible segmentations. 
\end{itemize}

The resulting size of the English dataset is 15k hashtags for training and 1k hashtags for testing.  
\vspace{-1mm}
\begin{table}[h!]
\centering
\caption{Samples from both datasets}
\label{tbl4}
\begin{tabular}{|l|m{3.2cm}|m{2.8cm}|} \hline 
	& Russian dataset & English dataset \\ \hline 
    \begin{turn}{90} Train \end{turn} & \makecell{мвдпетровкадети \\ ПоисковыхРабот09 \\ КМСстихи \\ середины\_века \\ Будем\_Рады } & \makecell{sunwouldcome \\ ThingsGoingWell \\ StartSchool\_72 \\ tonightgoodnight \\ muchloveu} \\ \hline
	\begin{turn}{90} Test \end{turn} & \makecell{ЛайфхакМЧС \\ важно\_знать \\ ЗавтраБылаВойна \\ важнаядата \\ ФИФА2018} & \makecell{KrispyKreme \\ twitteriffic \\ titsuptuesday \\ MissUSA \\ ipv6summit} \\ \hline
\end{tabular}
\end{table}

\section{Active Learning}
We followed the strategy for active learning, as in \cite{shen2017deep}.  The training procedure consists of multiple rounds of training and testing of the model.  We start by training the model on 1k hashtags, which were randomly selected from the training dataset. Next we test the model on the reminder of the training dataset and select 1k hashtags according to the current model’s uncertainty in its prediction of the segmentation. These hashtags are not manually relabelled, since a) they belong to the synthetically generated training dataset and b) the correct labeling for these hashtag is already known.  In  \cite{shen2017deep} three uncertainty measure are presented, from which we selected the maximum normalized log-probability (MNLP) assigned by the model to the most likely sequence of tags. The model is then retrained on the hashtags it is uncertain about. Note, that here we do not check if the predictions of the model are correct. We are more interested in training the model on hard examples than in evaluating the quality of intermediate results. We refer the reader to \cite{shen2017deep} for more technical details.

\section{Experiments}
\subsection{Baseline}

As for baseline algorithm, we consider the \cite{bansal2015towards} system architecture as a state-of-the-art algorithm. Unfortunately, their approach is not straightforwardly applicable to our synthetic Russian dataset, because it requires twofold input: a hashtag and a corresponding tweet or a text from any other social media, which is absent in our task setting due to synthetic nature of the training dataset.

For this reason as a baseline algorithm for English dataset we refer to results from \cite{bansal2015towards}, and as for Russian dataset, we used the probabilistic language model, described in \cite{norvig2009natural}. The probability of a sequence of words is the product of the probabilities of each word, given the word’s context: the preceding word. As in the following equation:

\[p(w_1, w_2, .., w_n) =  \prod_{i = 1}^{n}  p(w_i \vert w_{i-1}), \] 
where 
	\[ p(w_i \vert w_{i-1}) = \frac{f(w_{i-1}, w_i)}{f(w_{i-1})} \]

In case there is no such a pair of words $(w_{i-1}, w_i)$ in the set of bigrams, the probability of word $w_i$ is obtained as if it was only an unigram model:


\begin{align*}
	& p(w_i) = \frac{f(w_i) + \alpha} {\sum_{j=1}^ {\vert W \vert} f(w_j)  + \alpha \vert V \vert }
\end{align*}

where $V$ – vocabulary, $f(w_{i})$ – frequency of word $w_{i}$, and $\alpha$ = 1.


In Table \ref{tbl5} we present three baseline results: LM \cite{norvig2009natural} for Russian and English datasets; context-based LM \cite{bansal2015towards} for English dataset only. We treat a segmentation as correct if prediction and target sequences are the same.
\vspace{-1mm}
\begin{table}[h!]
\centering
\caption{Accuracy of the baseline algorithm on the Russian and English datasets}
\label{tbl5}
\begin{tabular}{|l|l|} \hline 
& Accuracy \\ \hline 
Russian dataset \cite{norvig2009natural} & 0.634 \\ \hline 
English dataset \cite{norvig2009natural} & 0.526 \\ \hline
English dataset \cite{bansal2015towards} & 0.711 \\ \hline
\end{tabular}
\vspace{-1mm}
\end{table}

\subsection{Neural Model}

In our experiments we used 5 epochs to train the char-RNN with LSTM units. For each language we observed three datasets with different number of hashtags. In case of Russian language, the more data we use while training, the higher the accuracy. As for English, the highest accuracy score was achieved on a set of 10k hashtags (Table \ref{tbl6}). Due to it's lower morphological diversity and complexity the model starts to overfit on training sets with large sizes. The training showed that mostly the model makes wrong predictions of segmentation on hashtags of complex types, such as ``wordword\_worddigits''.


\begin{table}[h!]
\centering
\caption{Accuracy of LSTM char-RNN on both Russian and English datasets}
\label{tbl6}
\begin{tabular}{|l|m{3cm}|m{3cm}|} \hline 
 & Accuracy, Russian dataset & Accuracy, English dataset \\ \hline 
5k & 0.9682 & 0.9088 \\ \hline
10k & 0.9765 & \textbf{0.9134} \\ \hline
15k & \textbf{0.9811} & 0.8733 \\ \hline
\end{tabular}
\vspace{-1mm}
\end{table}

Our results outperform all choosen baseline both for Russian and English datasets. Note, that we have two baselines for the English dataset: one is purely frequency-based, another is cited from \cite{bansal2015towards}, where external resources are heavily used. We show that using significantly less amount of training data, we achieve a boost in quality by switching from statistical word language models to char-RNN. 
As expected, the results on Russian dataset are higher than for the English dataset due to higher inflection degree in Russian as opposed to English. 

\subsection{Active Learning}

In order to evaluate the efficiency of deep learning with active learning when used in combination, we run the experiments for both languages. As for the datasets, we took the ones on which the highest accuracy was obtained (15k for Russian and 10k for English).

The learning process consists of multiple rounds which are repeated until the test set is finished. At the beginning we train the model on 1k of randomly selected hashtags and predict the probability of segmentation for the remaining hashtags. Then we sort the remaining hashtags in ascending order according to the probability assigned by the model and pick 1k of hashtags  which the model is least confident about. Finally, we add these hashtags with the least probable sequence of tags to the training data and continue training the model. This pipeline is repeated till there are no samples left.

In comparison to our initial experiments, application of active learning demonstrates impressive results. The amount of labeled training data can be drastically reduced, to be more specific, in both cases the size of the training set can be reduced by half without any decline in accuracy (see Figures 2 and 3).

Active learning  selects a more informative set of examples in contrast to supervised learning, which is trained on a set of randomly chosen examples. We decided to analyze the updated version of the training data and see if number of morphologically complex types of hashtags is higher than the simple ones. We were able to divide hashatgs into complex and simple as the model is trained on synthetic data and there is a finite number of templates by which each hashtag can be generated.

\begin{figure}[h!]
\centering
\caption{Accuracy obtained on Russian Dataset}
\includegraphics[width = 0.7\textwidth]{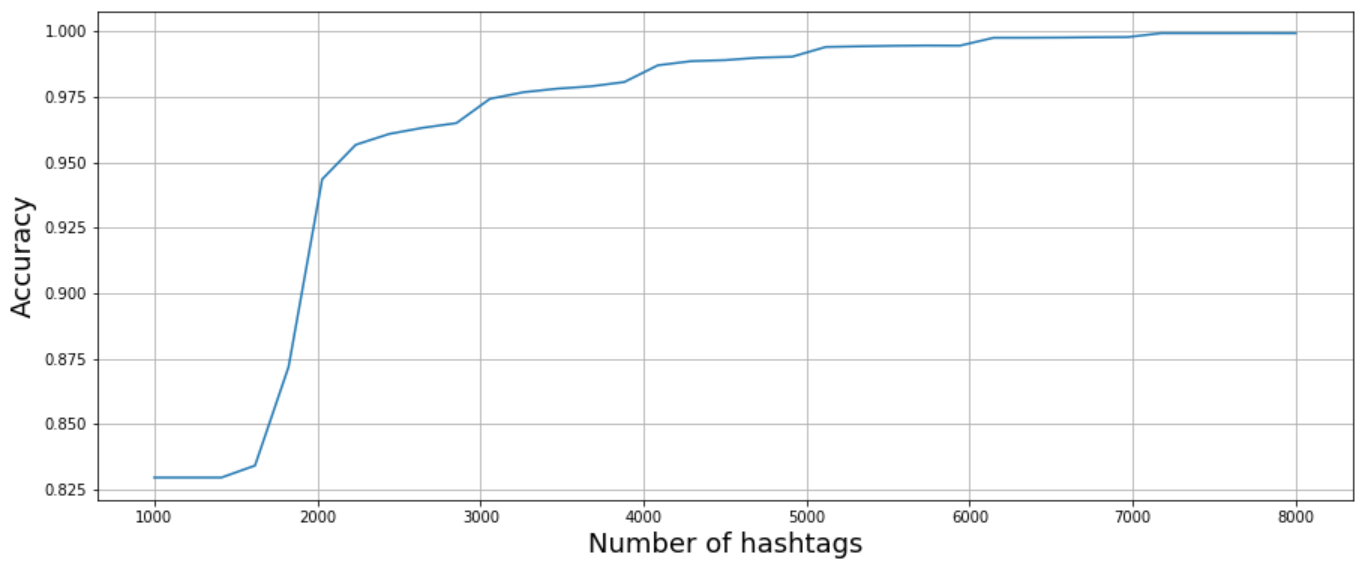}

\centering
\caption{Accuracy obtained on English Dataset}
\includegraphics[width = 0.7\textwidth]{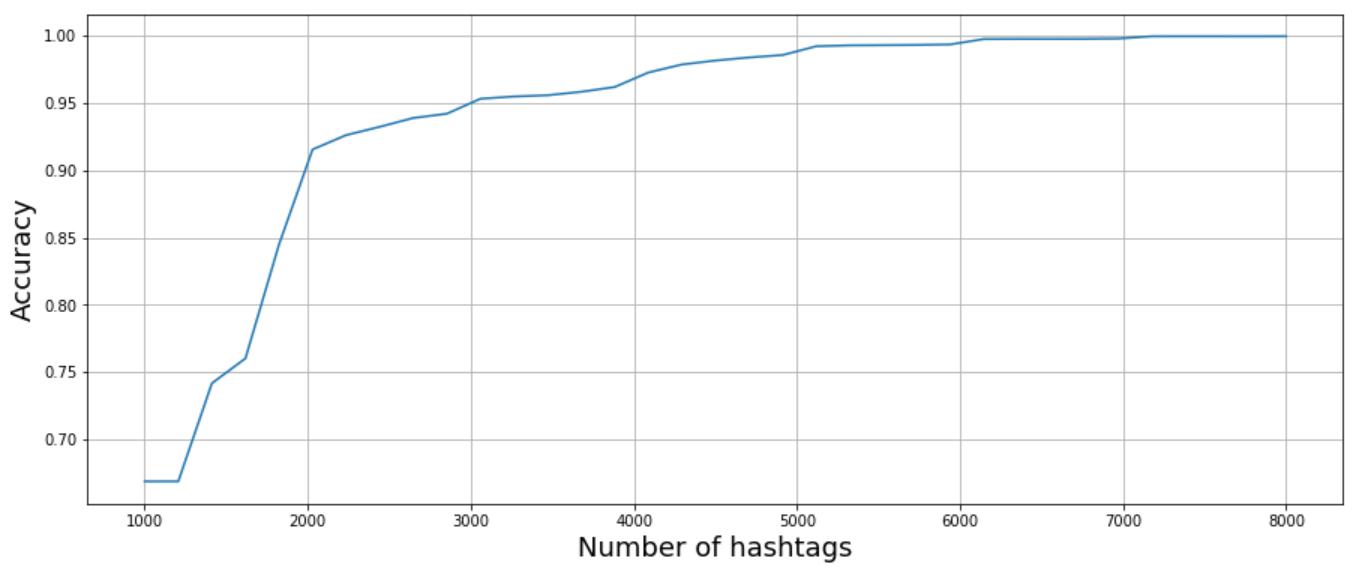}
\end{figure}

To better understand the contribution of uncertainty sampling approach, we plot the distribution of different types of hashtags in new training datasets for both languages, Russian and English (see Figure 4 and 5). According to identified types of hashtags in real data, it can be seen from the plots that in both cases the algorithm added more of morphologically complex hashtags to training data – types 3, 6 and 7. These types mostly consist of hashtags with two or three words in lower case without underscore.

Examples of featured types:

\begin{description}
	\item [type 3:] wordword\_2017
    \item [type 6:] wordword, word2017word
    \item [type 7:] wordwordword, wordword2017word
\end{description}

\begin{figure}[h!]
\centering
\caption{Distribution of top 7k hashtag types in Russian dataset, chosen by an active learning algorithm}
\includegraphics[width = 0.7\textwidth]{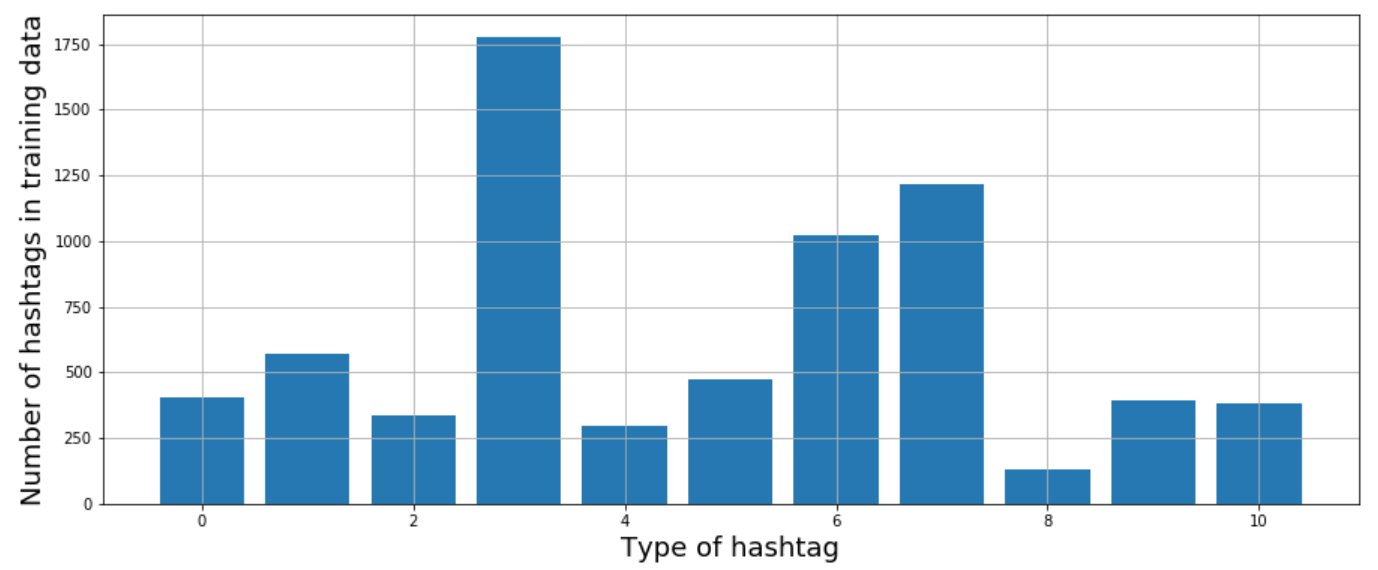}

\centering
\caption{Distribution of top 7k hashtag types in English dataset, chosen by an active learning algorithm}
\includegraphics[width = 0.7\textwidth]{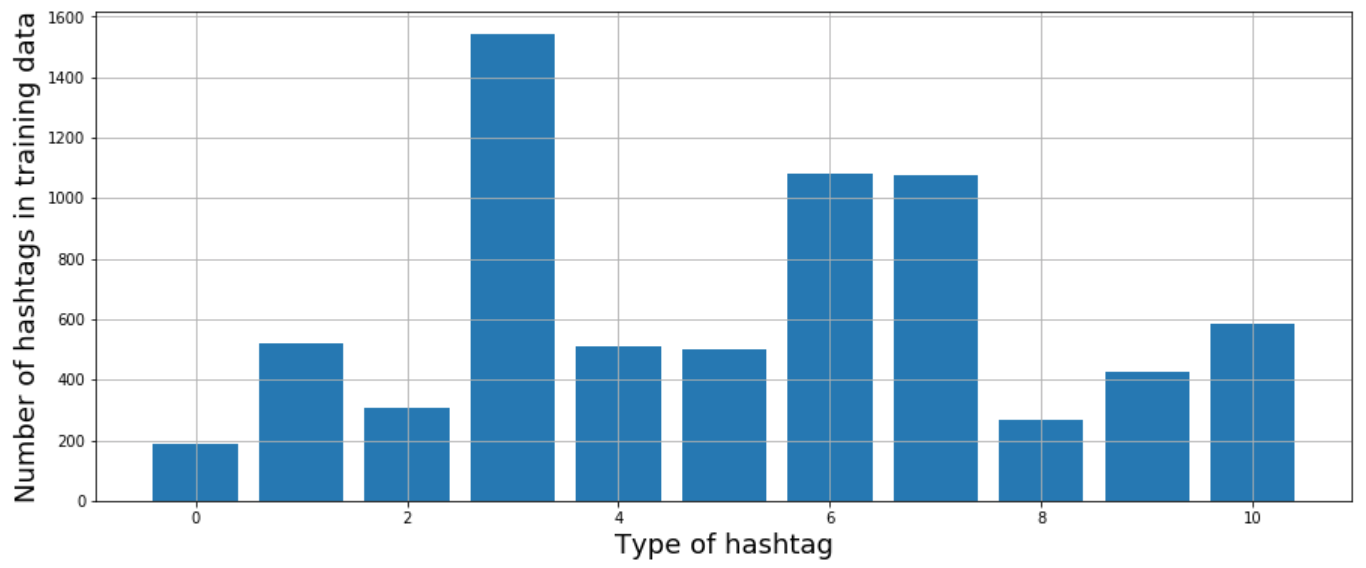}
\end{figure}

\subsection{Visualization}

In order to see if embeddings of similar characters, in terms of string segmentation, appear near each-other in their resulting 50-dimensional embedding space, we have applied one technique for dimensionality reduction: SVD to character embeddings to plot them on 2D space. For both languages meaningful and interpretable clusters can be extracted: capital letters, letters in lower case, digits and underscore, as shown below.

\begin{figure}[h!]
\centering
\caption{SVD visualization of character embeddings on Russian dataset}
\includegraphics[width = 0.7\textwidth]{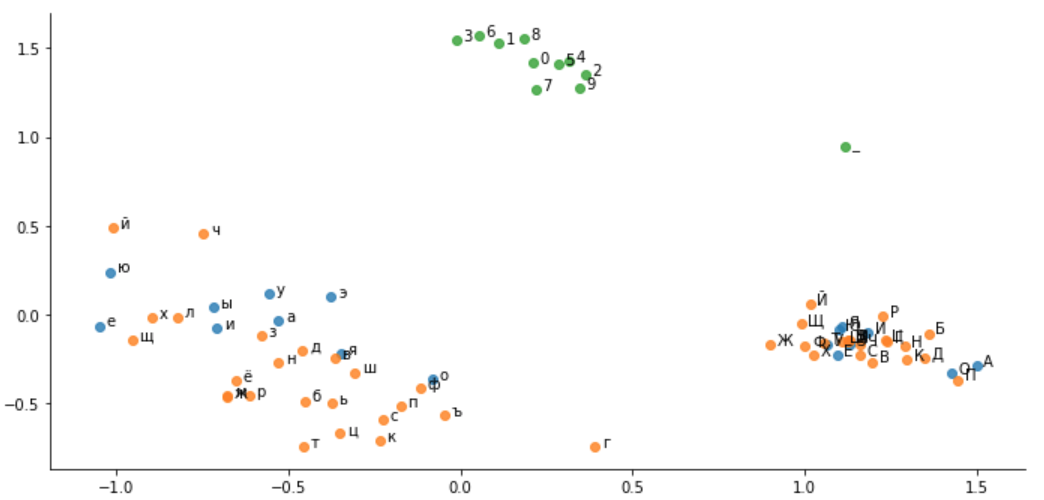}
\end{figure}

\begin{figure}[h!]
\centering
\caption{SVD visualization of character embeddings on English dataset}
\includegraphics[width = 0.7\textwidth]{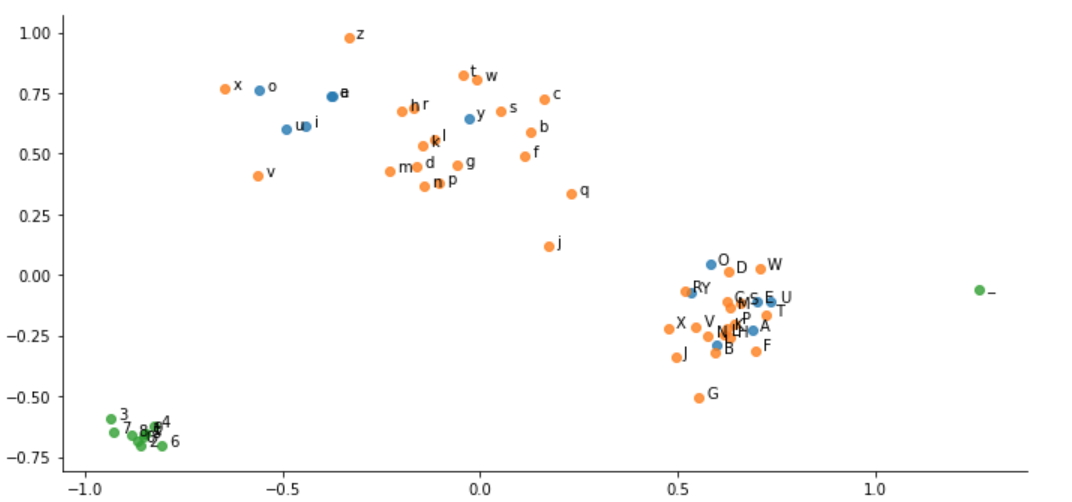}
\end{figure}

\section{Related Work}

The problem of word segmentation has received much attention in Chinese and German NLP for word segmentation and compound splitting \cite{xue2003chinese}, respectively. The major techniques for word segmentation exploit string matching algorithms \cite{reuter2016segmenting}, language models  \cite{berardi2011isti,bansal2015towards} and sequence labeling methods \cite{xue2003chinese}. Recent trend of deep learning as a major approach for any NLP task in general and sequence labeling in particular resulted in using various RNN-based models and CNN-based model for Chinese word segmentation \cite{xue2003chinese,koehn2003empirical,riedl2016unsupervised}.

Since \cite{xue2003chinese} Chinese word segmentation is addressed as a character labeling task: each character of the input sequence is labeled with one of the four labels $\mathcal{L} = \{B, M, E, S\}$, which stand for character in Begin, Middle or End of the word or Single character word. \cite{xue2003chinese} uses a maximum entropy tagger to tag each character independently. This approach was extended in \cite{peng2004chinese} to the sequence modeling task, and linear conditional random fields were used to attempt it and receive state of the art results. A neural approach to Chinese segmentation mainly uses various architectures of character level recurrent neural networks \cite{cai2016neural,zhang2018neural,cai2017fast} and very deep constitutional networks \cite{sun2017gap}. Same architectures are used for dialectal Arabic segmentation \cite{samih2017neural}.

The evolution of German compound splitters is more or less similar to Chinese word segmentation systems. The studies of German compound splitting started with corpus- and frequency-based approaches \cite{koehn2003empirical,riedl2016unsupervised} and are now dominated with neural-based distributional semantic models. However, German compound splitting is rarely seen as sequence modeling task. 

The problem of hashtag segmentation, analysis and usage in English has been approached by several research groups. As it was shown by  \cite{berardi2011isti} hashtag segmentation for TREC microblog track 2011 \cite{ounis2011overview} improves the quality of information retrieval, while \cite{bansal2015towards} shows that hashtag segmentation improves linking of entities extracted from tweets to a knowledge base. Both \cite{berardi2011isti,bansal2015towards} use Viterbi-like algorithm for hashtag segmentation: all possible segmentations of hashtag are scored using a scoring function:

\[ \texttt{Score}(S) =  \sum_{s_i \in S} \log (P_{Unigram}(s_i)), \]

where $P_{Unigram}$ are probabilities, computed according to the unigram model based on a large enough corpus or any N-gram service. 

Following the idea of scoring segmentation candidates, \cite{reuter2016segmenting} introduces other scoring functions, which include a bigram model (2GM) and a Maximum Unknown Matching (MUM), which is adjustable to unseen words. 

\cite{declerck2015processing} attempt to split camel-cased hashtags using rule-based approach and POS-tagging for further semantic classification. WordSegment\footnote{\url{http://www.grantjenks.com/docs/wordsegment/}} has been used for sentiment analysis \cite{akhtar2017iitp,park2018plusemo2vec} and other applications. 

To our knowledge there has been little work done for word or hashtag segmentation in Russian.

\subsection{Active Learning in NLP}

Active learning is machine learning technique which allows efficient use of the available training data. It presumes that, first an initial model is trained on a very little amount of data and next tested on large unlabeled set. Next the model is able to choose a few most difficult examples and ask an external knowledge source about the desired labels. Upon receiving these labels, the model is updated and retrained on the new train set. There might be a few rounds of label querying and model updating. To use active learning strategy, we need a definition of what a difficult example is and how to score its difficulty. One of the most common scoring approaches is  entropy-based uncertainty sampling, which selects the examples with the lowest prediction probability. 

Active learning is widely used in NLP applications, when there is little annotated data while the amount of unlabeled data is abundant. Being ultimately used for text classification using traditional machine learning classifiers \cite{tong2001support,schohn2000less}, active learning is less known to be used with deep learning sequence classifiers. Recent works report on scoring word embeddings that are likely to be updated with the greatest magnitude \cite{zhang2017active} and on using maximum normalized log-probability (MNLP) assigned by the model to the most likely sequence of tags \cite{shen2017deep}:

\[ \frac{1}{n} \max_{y_1 \ldots y_n} \sum_{i=1}^n \log P[y_n | y_1, \ldots, y_{n-1} , x_{ij}] \]

\subsection{Training on synthetic data}
The lack of training data is an issue for many NLP applications. There have been attempts to generate and use synthetic data for training question answering systems \cite{weston2015towards} and SQL2text systems \cite{utama2018end}. In \cite{bansal2015towards} synthetic hashtags are generated by removing whitespace characters from frequent n-grams, while in \cite{matthews2016synthesizing} German compounds are synthesized for further machine translation.  



\section{Conclusions}
In this paper we approach the problem of hashtag segmentation by using char-RNNs. We treat the problem of hashtag segmentation as a sequence labeling task, so that each symbol of a given string is labeled with 1 (there should be a whitespace after this symbol) or 0 (otherwise). We use two datasets to test this approach in English and in Russian without any language-specific settings. We compare char-RNN to traditional probabilistic algorithms. To interpret the results we use a few visualization techniques and the strategy of active learning to evaluate the complexity of training data, since we use synthetically generated hashtags for training.  

The results show that:

\begin{enumerate}
\item When approached on character level, hashtag segmentation problem can be solved using relatively small and simple recurrent neural network model without usage of any external corpora and vocabularies. Such char-RNN not only outperforms significantly traditional frequency-based language models, but also can be trained on synthetic data generated according to morpho-syntactic patterns, without any manual annotation and preprocessing.

\item In languages with high inflection (such as Russian) the char-RNN achieves higher results than in languages with little inflections (such as English) due to the ability of the char-RNN to capture and memorize word boundary patterns, especially word ending patterns (i.e. adjective endings ``ый'',``ая'',``ое'' or verbal endings ``ать'',``еть'' in Russian). 

\item The amount of generated synthetic training data can be limited by using techniques for active learning which allows to select sufficient training subset without any loss of quality.

\end{enumerate}

\section{Acknowledgements}

The paper was prepared within the framework of the HSE University Basic Research Program and funded by the Russian Academic Excellence Project '5-100'. 

%
%
%
%

\end{document}